\documentclass[10pt,twocolumn,letterpaper]{article}

\usepackage[pagenumbers]{cvpr} 

\usepackage{graphicx}
\usepackage{amsmath}
\usepackage{amssymb}
\usepackage{booktabs}
\usepackage{multirow}
\usepackage{comment}
\usepackage{caption}
\usepackage{wrapfig}
\usepackage{enumitem}
\usepackage{mathtools}
\usepackage{array}
\usepackage{color}
\usepackage{algorithm}
\usepackage{colortbl}
\definecolor{greyline}{rgb}{0.105,0.410,0.113}
\newcommand{\squishlist}{
	\begin{list}{$\bullet$}
		{ \setlength{\itemsep}{0pt}
			\setlength{\parsep}{1pt}
			\setlength{\topsep}{1pt}
			\setlength{\partopsep}{0pt}
			\setlength{\leftmargin}{1.5em}
			\setlength{\labelwidth}{1em}
			\setlength{\labelsep}{0.5em} } }
\newcommand{\squishend}{\end{list} 
}
\usepackage[most]{tcolorbox}
\usepackage{tcolorbox}

\usepackage[toc,page]{appendix}

\definecolor{navyblue}{HTML}{0071BC}
\definecolor{hotpink}{HTML}{FF0080}

\definecolor{oai-white}{HTML}{FFFFFF}
\definecolor{oai-black}{HTML}{000000}
\definecolor{oai-red}{HTML}{FF4500}
\definecolor{oai-green}{HTML}{51DA4C}
\definecolor{oai-blue}{HTML}{0000FF}
\definecolor{oai-yellow}{HTML}{FFF639}
\definecolor{oai-magenta}{HTML}{FF45FF}
\definecolor{oai-cyan}{HTML}{00FFFF}
\definecolor{oai-orange}{HTML}{FE7600}
\definecolor{oai-violet}{HTML}{8A2BE2}
\definecolor{oai-brown}{HTML}{A0522D}
\definecolor{oai-green-050}{HTML}{F4FFF4}
\definecolor{oai-green-100}{HTML}{E9FFE8}
\definecolor{oai-green-200}{HTML}{D9FFD8}
\definecolor{oai-green-300}{HTML}{C9FFC7}
\definecolor{oai-green-400}{HTML}{A6FFA3}
\definecolor{oai-green-500}{HTML}{7CF178}
\definecolor{oai-green-600}{HTML}{51DA4C}
\definecolor{oai-green-700}{HTML}{3FA93B}
\definecolor{oai-green-800}{HTML}{2D712A}
\definecolor{oai-green-900}{HTML}{193718}
\definecolor{oai-gray-000}{HTML}{FFFFFF}
\definecolor{oai-gray-100}{HTML}{FAFAFA}
\definecolor{oai-gray-200}{HTML}{F5F5F5}
\definecolor{oai-gray-300}{HTML}{E5E5E5}
\definecolor{oai-gray-400}{HTML}{FFB7A4}
\definecolor{oai-gray-500}{HTML}{CDCDCD}
\definecolor{oai-gray-600}{HTML}{A8A8A8}
\definecolor{oai-gray-700}{HTML}{747474}
\definecolor{oai-gray-800}{HTML}{393939}
\definecolor{oai-gray-900}{HTML}{000000}
\definecolor{visual}{HTML}{A50E0E}       
\definecolor{linguistic}{HTML}{174EA6}   
\definecolor{relational}{HTML}{E37400}   
\definecolor{egocentric}{HTML}{0D652D}  
\colorlet{mapcolor}{ForestGreen}

\newcommand{\infobox}[1]{
    \vspace{-0.18cm}
    \begin{tcolorbox}[
        colback=white!90!gray,     
        colframe=teal!60!black,   
        arc=5pt,                   
        boxsep=5pt,                 
        left=5pt,                  
        right=10pt,                 
        top=2pt,                   
        bottom=3pt,                
        boxrule=0.8pt,              
        drop shadow=gray!50!white, 
        enhanced jigsaw             
    ]
    \vspace{-0.1cm}
         \textit{#1}
    \vspace{-0.2cm}
    \end{tcolorbox}
    \vspace{-0.15cm}
}

\newcommand{\remarkbox}[1]{
    \vspace{-0.18cm}
    \begin{tcolorbox}[
        colback=blue!5!white,
        colframe=blue!70!black,
        arc=5pt,
        boxsep=5pt,
        left=5pt,
        right=10pt,
        top=2pt,
        bottom=3pt,
        boxrule=0.8pt,
        drop shadow=gray!40!white,
        enhanced jigsaw
    ]
    \vspace{-0.1cm}
         \textit{#1}
    \vspace{-0.2cm}
    \end{tcolorbox}
    \vspace{-0.15cm}
}

\definecolor{cvprblue}{rgb}{0.21,0.49,0.74}
\usepackage[pagebackref,breaklinks,colorlinks,allcolors=cvprblue]{hyperref}

\usepackage{xcolor}
\definecolor{visualgreen}{HTML}{4FAD5B}
\definecolor{visualpurple}{HTML}{68349A}
\definecolor{visualred}{HTML}{B22E25}
\definecolor{visualblue}{HTML}{4270B1}

\usepackage[capitalize]{cleveref}
\crefname{section}{Sec.}{Secs.}
\Crefname{section}{Section}{Sections}
\Crefname{table}{Table}{Tables}
\crefname{table}{Tab.}{Tabs.}

\def\paperID{***} 
\def\confName{CVPR}
\def\confYear{2026}

\title{Rethinking Chain-of-Thought Reasoning for Videos}

\author{Yiwu Zhong$^{1}$, Zi-Yuan Hu$^{1}$, Yin Li$^{2}$, Liwei Wang$^{1}$\thanks{Corresponding author.}\\
$^1$The Chinese University of Hong Kong, $^2$University of Wisconsin-Madison\\
}

\begin{document}
\maketitle

\begin{abstract}
Chain-of-thought (CoT) reasoning has been highly successful in solving complex tasks in natural language processing, and recent multimodal large language models (MLLMs) have extended this paradigm to video reasoning. However, these models typically build on lengthy reasoning chains and large numbers of input visual tokens. Motivated by empirical observations from our benchmark study, we hypothesize that concise reasoning combined with a reduced set of visual tokens can be sufficient for effective video reasoning. To evaluate this hypothesis, we design and validate an efficient post-training and inference framework that enhances a video MLLM's reasoning capability. Our framework enables models to operate on compressed visual tokens and generate brief reasoning traces prior to answering. The resulting models achieve substantially improved inference efficiency, deliver competitive performance across diverse benchmarks, and avoid reliance on manual CoT annotations or supervised fine-tuning. Collectively, our results suggest that long, human-like CoT reasoning may not be necessary for general video reasoning, and that concise reasoning can be both effective and efficient. Our code will be released at \url{https://github.com/LaVi-Lab/Rethink_CoT_Video}.
\end{abstract}

\section{Introduction}
\label{sec:intro}

Chain-of-Thought (CoT)~\cite{cot2022,autocot2022} aims to solve complex tasks by generating explicit, step-by-step intermediate reasoning traces before producing a final answer. 
CoT has been a key driver of the strong reasoning capabilities in latest large language models (LLMs)~\cite{jaech2024openai,guo2025deepseek}. 
Building on this success, several recent efforts~\cite{video-r1,video-rft,llava-cot,vision-r1,r1-onevision} have extended CoT to multimodal large language models (MLLMs), demonstrating improved reasoning over visual inputs, including both images and videos.

Despite its success in vision, CoT reasoning in MLLMs incurs major overhead in both inference and training. 
On the inference side, visual inputs, especially videos, often expand into thousands of visual tokens with high redundancy, while CoT produces long reasoning sequences. Together, these factors compound to sharply increased memory usage, compute cost, and carbon footprint during deployment.
Additionally, CoT training relies on supervised fine-tuning (SFT) with heavily-labeled reasoning traces, followed by reinforcement learning (RL) over long sequences~\cite{video-r1,longvila-r1,video-rft, llava-cot,vision-r1,r1-onevision,r1-vl,reason-rft}. This pipeline not only requires costly annotation, but also leads to prolonged training cycles. 

\begin{figure}[t]
    \centering
    \includegraphics[width=0.9\linewidth]{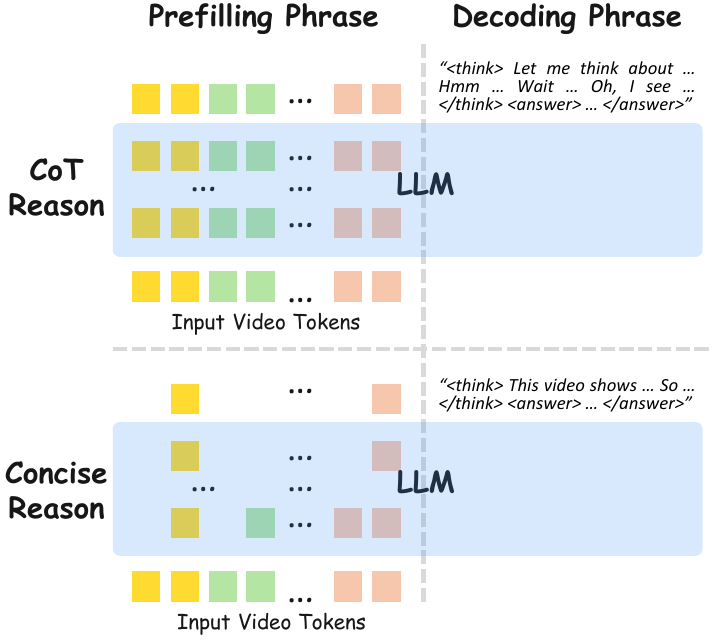}
    \vspace{-8pt}
    \caption{\textbf{CoT Reasoning}, with dense prefilling and lengthy decoding, incurs substatial computation load at both training and inference. In contrast, \textbf{Concise Reasoning} coupled with token compression is significantly more efficient, thanks to sparse prefilling and concise decoding.}
    \vspace{-10pt}
    \label{fig:teaser}
\end{figure}

In this paper, we aim to reduce the inference and training overhead of reasoning-oriented video MLLMs.
With a Transformer-based architecture (Fig.~\ref{fig:teaser}), this overhead can be decomposed into: 
(1) the cost of decoding phrase, which scales with the number of output tokens for reasoning and the total number of input tokens, 
and (2) the cost of prefilling phrase, which is determined by the number of input tokens.
Therefore, reducing CoT overhead requires either shortening the output chain, which promotes more concise reasoning, or decreasing the number of input tokens, which is dominated by redundant visual tokens and can be reduced via token compression~\cite{shang2024llava,zhong2024aim,zhang2024sparsevlm,tao2025dycoke,chen2024image}.
With this intuition, we conduct \textit{a systematic benchmark of MLLMs with and without CoT across a suite of video datasets}, covering general, long-form, and complex video understanding tasks.

Our benchmark leads to several surprising observations. 
\textbf{First}, despite the overhead, adding CoT yields only modest gains over the base pre-trained MLLM using direct answering. Indeed, we observe that CoT outputs frequently contain human-like ``pondering'' patterns (\eg, ``Hmm,'' ``Let’s think,'' or ``Wait'') that contribute little to reasoning.
\textbf{Second}, prompting the base pre-trained MLLM to generate concise reasoning chains leads to a major performance drop, significantly worse than direct answering. We conjecture that the model possesses the necessary knowledge to answer the questions, yet is not well aligned with the concise reasoning paradigm.
\textbf{Third}, we find that token compression, although previously shown to be effective for video MLLMs~\cite{zhong2024aim}, causes notably larger performance degradation when the model is prompted to produce concise or CoT-style reasoning, compared with direct answering.

Collectively, these observations motivate a rethinking of how CoT inference and training should be designed for video reasoning. While existing MLLMs with CoT typically rely on lengthy reasoning chains and a large number of input visual tokens, \textit{we hypothesize that concise reasoning combined with a reduced set of visual tokens can be sufficient for effective video reasoning}. To verify this, we design and validate an efficient post-training and inference framework that enhances video MLLM's reasoning capability. Specifically, we directly fine-tune a pre-trained video MLLM via RL using Group Relative Policy Optimization (GRPO)~\cite{shao2024deepseekmath}. We further integrate visual token compression with GRPO, allowing the model to learn to reason effectively from a reduced set of tokens. At inference time, the model operates on compressed visual tokens and produces brief reasoning traces before its final answer. In doing so, \textit{our framework substantially reduces inference overhead, demonstrates improved accuracy across a wide range of benchmarks, and additionally removes the need for costly CoT annotations and SFT training}. 

\smallskip
\noindent \textbf{Contributions.} 
Our benchmark, framework, and results challenge the prevailing assumption that long, human-like CoT reasoning is essential for video reasoning. Instead, we demonstrate an alternative approach that operates on compressed visual tokens and produces concise reasoning traces. The resulting models achieve improved inference efficiency, deliver competitive performance, and can be learned solely through RL-based post-training.

\smallskip
\noindent \textbf{Scope and limitations.} 
The objective of this work is not to introduce a new method, but rather to \textit{illuminate a parallel solution that deviates from a widespread view in the field}. We benchmark existing methods, draw empirical observations, formulate a hypothesis, and design a framework to test that hypothesis. In doing so, we do not claim algorithmic superiority over prior methods, in spite of better results. Our work is grounded in empirical evidence rather than theoretical guarantees, and our conclusions are constrained by the scope of current video benchmarks. Nonetheless, we believe the insights revealed in our study may generalize beyond these datasets and potentially extend to other forms of visual reasoning.

\section{Related Work}
\label{sec:related_work}

\noindent \textbf{Multi-Modal LLMs.}
Significant progress has been made in developing multi-modal LLMs for image understanding~\cite{dai2023instructblip,zhu2023minigpt4,liu2023improvedllava,lin2024vila}, video understanding~\cite{Maaz2023VideoChatGPT,lin2023video,zhu2023languagebind,mPLUG-Owl3,llavavideo}, and long-video reasoning~\cite{longvlm,longvila,video-xl,koala,moviechat+,MA-LLM}.
More recently, unified multimodal models~\cite{li2024llava,Qwen2VL,qwen25,qwen3,internvl3,liu2025nvila} have emerged, exhibiting strong generalization across both images and videos.
Building on these advances, we explore whether the concise reasoning capability learned through large-scale pre-training can be effectively elicited via reinforcement learning for general video understanding, without relying on additional CoT annotations.

\smallskip
\noindent \textbf{Multi-modal Chain-of-Thought Reasoning.}
Inspired by CoT prompting in text-based LLMs~\cite{wei2022chain,kojima2022large}, early efforts in multi-modal reasoning focus on generating intermediate reasoning steps for visual understanding~\cite{zhong-2024-beyond,CCoT2024,mmcot2023,ddcot2024, VoT2024,Visualcot2024,step2025,videoespresso2025}.
Recent breakthroughs such as OpenAI-o1~\cite{jaech2024openai} and DeepSeek-R1~\cite{guo2025deepseek} demonstrate that reinforcement learning algorithms (\eg, RLHF~\cite{RLHF2022}, DPO~\cite{dpo2023}, and GRPO~\cite{shao2024deepseekmath}) can significantly enhance reasoning capability.
Among them, GRPO, featuring rule-based rewards and group-based sampling, has become a scalable framework for post-training and has been widely extended to vision tasks~\cite{visual-rft,vlm-r1} and vision-language tasks, including image understanding~\cite{insight-v,llava-cot,vision-r1,r1-onevision,visionary-r1,r1-vl,reason-rft, vl-rethinker,pixel-reasoner} and video reasoning~\cite{video-r1,longvila-r1,video-rft,ego-r1,videochat-r15,videochat-r1}.

A common practice in these methods is a two-stage post-training pipeline~\cite{video-r1,longvila-r1,video-rft,ego-r1, llava-cot,vision-r1,r1-onevision,r1-vl,reason-rft, pixel-reasoner, hybrid-think}: first generating CoT annotations for supervised fine-tuning (SFT), followed by GRPO-based reinforcement learning (RL). 
Specifically, Video-R1~\cite{video-r1} designs temporal-aware reward for GRPO fine-tuning and Video-RFT~\cite{video-rft} proposes a semantic consistency reward to align CoT generation and video descriptions. LongVILA-R1~\cite{longvila-r1} builds sequence parallelism for long videos and Ego-R1\cite{ego-r1} trains a tool-using agent for ego-centric long videos. VideoChat-R1.5~\cite{videochat-r15} learns to iteratively refine the focused regions by the collected spatio-temporal supervision. 
In this work, we challenge the prevailing paradigm of CoT reasoning and its two-stage SFT-GRPO fine-tuning. Our experiments suggest that direct GRPO training with concise reasoning is already sufficient to elicit strong multi-modal reasoning, removing the heavy overhead for CoT annotation or CoT SFT.

\smallskip
\noindent \textbf{Token Compression.} 
Token compression methods aim to reduce the number of tokens processed by Transformer-based models, thereby lowering computational overhead. Such techniques have been explored in NLP models~\cite{powerbert,lat,spatten,trbert,learned}, vision models~\cite{dynamicvit,avit,spvit,tps,savit,dtop,heatvit, ats,bolya2022tome,wang2024zero}, and multi-modal LLMs~\cite{chen2024image,lin2024boosting,xing2024pyramiddrop,shang2024llava,zhong2024aim,zhang2024treat,ye2025fit,zhang2024sparsevlm,tao2025dycoke}.
Among them, several approaches operate without re-training and can be directly applied during inference~\cite{zhang2024sparsevlm,chen2024image,zhong2024aim,tao2025dycoke}. For instance, AIM~\cite{zhong2024aim} merges visually similar tokens before the LLM and prunes uninformative tokens within the LLM, achieving substantial token reduction with minimal impact on pre-trained capability.
Building on these characteristics, we incorporate token compression to improve the efficiency of both training and inference, particularly for processing long videos.

\section{Benchmarks and Analysis}
\label{sec:analysis}

In this section, we provide analysis on prevalent pre-trained models and typical chain-of-thought models, through accuracy and efficiency evaluation across diverse video benchmarks. These analysis motivate our final model design.

\begin{figure}[t]
    \centering
    \includegraphics[width=0.75\linewidth]{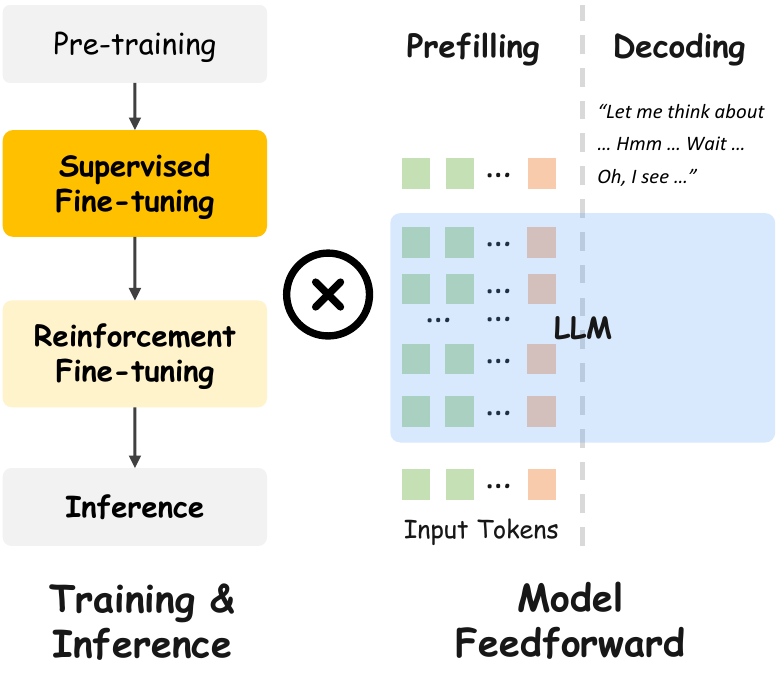}
    \vspace{-8pt}
    \caption{\textbf{Overview of CoT models}. After pre-training, they are typically post-trained via SFT stage using CoT annotations and RL stage. For both training and inference, the models suffer from heavy prefilling with dense visual tokens, and lengthy decoding due to human-like thinking generation.}
    \vspace{-10pt}
    \label{fig:analysis}
\end{figure}

\begin{figure}[t]
    \centering
    \includegraphics[width=0.99\linewidth]{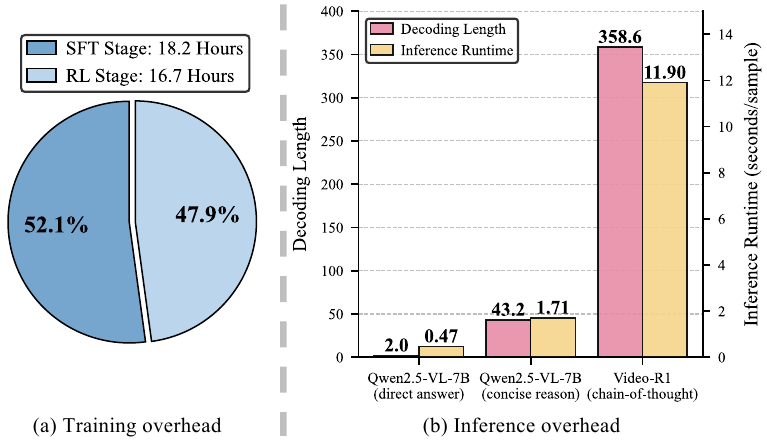}
    \vspace{-8pt}
    \caption{\textbf{Statistics of training and inference overhead}. 
    (a) Training overhead shows the training runtime of a CoT model (\ie, Video-R1~\cite{video-r1}), which is measured via four A800-SXM4-80GB GPUs.
    (b) Inference overhead reports the inference statistics (\ie, decoding length and inference runtime) of different reason modes, which is measured through a single A800-SXM4-80GB GPU.}
    \vspace{-10pt}
    \label{fig:overhead}
\end{figure}

\begin{table*}[t!]
\centering
\setlength{\tabcolsep}{8pt}
\renewcommand{\arraystretch}{1.4}
\resizebox{0.99\textwidth}{!}
{%
\begin{tabular}{@{}l|c|cc|cccc|ccc@{}}
    \toprule
    \multirow{2}{*}{\textbf{Model}} & 
    \multirow{2}{*}{\textbf{Decoding Mode}} &
    \multicolumn{2}{c|}{\textbf{\footnotesize{General Video Benchmarks}}} & 
    \multicolumn{4}{c|}{\textbf{\footnotesize{Long Video Benchmarks}}} & 
    \multicolumn{3}{c}{\textbf{\footnotesize{Complex Video Benchmarks}}} \\ 
    \cmidrule(l){3-11} 
     &  & {\textbf{\footnotesize{VideoMME}}} & {\textbf{\footnotesize{MVBench}}} & {\textbf{\footnotesize{MLVU}}} & {\textbf{\footnotesize{LVBench}}} & {\textbf{\footnotesize{LongVideoBench}}} & {\textbf{\footnotesize{EgoSchema}}} & {\textbf{\footnotesize{VideoHolmes}}} & {\textbf{\footnotesize{Video-TT}}} &  {\textbf{\footnotesize{MMVU}}} \\ \midrule
    Qwen2.5-VL-7B~\cite{qwen25} & direct answer &  55.5 & 63.2 & 55.4  & 34.2  & 52.5  &  52.5 & 35.7 & 35.2 & 63.4 \\
    Qwen2.5-VL-7B + Video-R1~\cite{video-r1} & chain-of-thought & 54.9 (\textcolor{oai-magenta}{-0.6}) & 64.9 (\textcolor{oai-green}{+1.7}) & 58.9 (\textcolor{oai-green}{+3.5}) & 35.4 (\textcolor{oai-green}{+1.2}) & 54.6 (\textcolor{oai-green}{+2.1}) & 47.6 (\textcolor{oai-magenta}{-4.9}) & 39.4 (\textcolor{oai-green}{+3.7}) & 39.9 (\textcolor{oai-green}{+4.7}) & 62.4 (\textcolor{oai-magenta}{-1.0}) \\ \midrule
    Qwen2.5-VL-7B~\cite{qwen25} & concise reason &  52.8 & 55.2 & 47.3  & 32.2  & 47.5  &  53.3 & 32.1 & 31.4 & 55.2 \\
    \bottomrule
    \end{tabular}%
}
\vspace{-6pt}
\caption{\textbf{Inference results on video benchmarks}. The CoT model improves the pre-trained model on some benchmarks yet not the others. This improvement requires substantial computations, including: CoT annotations, supervised fine-tuning, and lengthy decoding.}
\label{tab:inference_results}
\vspace{-6pt}
\end{table*}

\begin{table*}[t!]
\centering
\setlength{\tabcolsep}{8pt}
\renewcommand{\arraystretch}{1.4}
\resizebox{0.99\textwidth}{!}
{%
\begin{tabular}{@{}l|c|cc|cccc|ccc|c@{}}
    \toprule
    \multirow{2}{*}{\textbf{Model}} & 
    \textbf{Decoding} &
    \multicolumn{2}{c|}{\textbf{\footnotesize{General Video Benchmarks}}} & 
    \multicolumn{4}{c|}{\textbf{\footnotesize{Long Video Benchmarks}}} & 
    \multicolumn{3}{c|}{\textbf{\footnotesize{Complex Video Benchmarks}}} & \multirow{2}{*}{\textbf{Avg.}} \\ 
    \cmidrule(l){3-11} 
     & \textbf{Mode} & {\textbf{\footnotesize{VideoMME}}} & {\textbf{\footnotesize{MVBench}}} & {\textbf{\footnotesize{MLVU}}} & {\textbf{\footnotesize{LVBench}}} & {\textbf{\footnotesize{LongVideoBench}}} & {\textbf{\footnotesize{EgoSchema}}} & {\textbf{\footnotesize{VideoHolmes}}} & {\textbf{\footnotesize{Video-TT}}} &  {\textbf{\footnotesize{MMVU}}} & \\ \midrule
    Qwen2.5-VL-7B~\cite{qwen25} & direct answer &  52.5 (\textcolor{oai-magenta}{-3.0}) & 60.0 (\textcolor{oai-magenta}{-3.2}) & 54.6 (\textcolor{oai-magenta}{-0.8}) & 31.3 (\textcolor{oai-magenta}{-2.9}) & 51.1 (\textcolor{oai-magenta}{-1.4}) & 51.3 (\textcolor{oai-magenta}{-1.2}) & 34.7 (\textcolor{oai-magenta}{-1.0}) & 31.2 (\textcolor{oai-magenta}{-4.0}) & 63.8 (\textcolor{oai-magenta}{-0.4}) & \textcolor{oai-magenta}{\textbf{-1.9}} \\
    Qwen2.5-VL-7B + Video-R1~\cite{video-r1} & chain-of-thought & 52.8 (\textcolor{oai-magenta}{-2.1}) &  60.0 (\textcolor{oai-magenta}{-4.9}) & 57.4 (\textcolor{oai-magenta}{-1.5})  & 32.5 (\textcolor{oai-magenta}{-2.9}) &  50.4 (\textcolor{oai-magenta}{-4.2}) & 45.5 (\textcolor{oai-magenta}{-2.1}) & 37.8 (\textcolor{oai-magenta}{-1.6}) & 38.0 (\textcolor{oai-magenta}{-1.9}) & 61.8 (\textcolor{oai-magenta}{-0.6}) & \textcolor{oai-magenta}{\textbf{-2.4}}\\ \midrule
    Qwen2.5-VL-7B~\cite{qwen25} & concise reason &  46.7 (\textcolor{oai-magenta}{-6.1}) & 52.0 (\textcolor{oai-magenta}{-3.2}) & 39.4 (\textcolor{oai-magenta}{-7.9})  &  29.5 (\textcolor{oai-magenta}{-2.7}) & 44.4 (\textcolor{oai-magenta}{-3.1}) & 50.7 (\textcolor{oai-magenta}{-2.6}) & 26.9 (\textcolor{oai-magenta}{-5.2}) & 30.2 (\textcolor{oai-magenta}{-1.2}) & 48.2 (\textcolor{oai-magenta}{-7.0}) & \textcolor{oai-magenta}{\textbf{-4.3}} \\
    \bottomrule
    \end{tabular}%
}
\vspace{-6pt}
\caption{\textbf{Training-free token compression method applied during inference~\cite{zhong2024aim}}. The accuracy delta (colored) represents the performance degradation compared to the model without using token compression. 
Concise reason is largely impacted by token compression.
}
\label{tab:token_compression}
\vspace{-6pt}
\end{table*}

\begin{figure*}[t]
    \centering
    \includegraphics[width=0.9\linewidth]{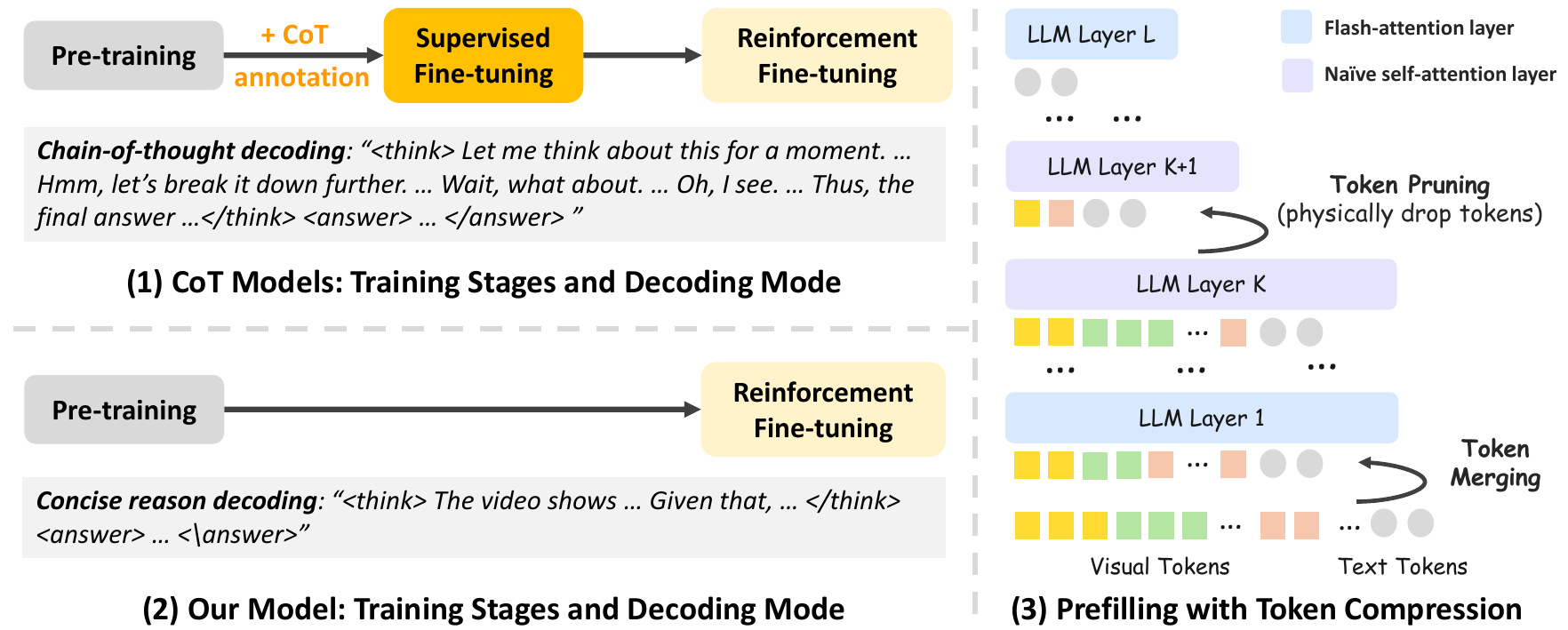}
    \vspace{-6pt}
    \caption{\textbf{Framework of our method}. (1) Typical CoT models are trained via three stages and perform long reasoning during inference. (2) In comparison, our method does not require the stage of supervised fine-tuning and the annotations at this stage, and generates concise reasoning during inference. (3) We further reduce computation overhead by trainable token compression.}
    \vspace{-8pt}
    \label{fig:method}
\end{figure*}

\subsection{Background}
\label{sec:background}

Before the anlysis, we first provide the background of our analysis as shown in Figure~\ref{fig:analysis}.

\smallskip
\noindent \textbf{Model training}: MLLMs are usually pre-trained on large-scale image-text and video-text datasets. 
Building on the capabilities acquired during pre-training, CoT models undergo a two-stage post-training process: 
(1) Supervised fine-tuning steers the models to learn the reasoning patterns derived from CoT annotations, 
and (2) Reinforcement fine-tuning seeks to incentivize reasoning capability through sparse reward signals and the GRPO algorithm~\cite{guo2025deepseek}.

\smallskip
\noindent \textbf{Model inference}: Both post-training and inference stages share the same model forward process, consisting of prefilling phrase and decoding phrase. 
In the prefilling phrase, the model processes input visual and text tokens to compute their key-value (KV) cashes at each LLM layer. The decoding phrase then auto-regressively generates tokens one by one. 
Both phrases are computationally impacted by the number of processed or generated tokens.

\smallskip
In the following, we describe more details on the different modes of decoding phrase.

\smallskip
\noindent \textbf{Chain-of-thought reason decoding}: CoT models are trained to generate long text traces that include multiple reasoning steps:
\begin{equation}
    y = \text{CoT}(x) = \{r_1, r_2, \dots, r_T, \hat{a}\},
\end{equation}
where $x$ denotes the input video and textual query, $y$ is the generated text response, $\{r_t\}$ denotes intermediate reasoning steps, and $\hat{a}$ is the final answer. 
These reasoning steps are typically human-like pondering patterns (\eg, ``Hmm,'', ``Let’s think,'', or ``Wait,'') which are learned from CoT annotations.
This decoding mode incurs high computational cost due to its lengthy token generation process.

\smallskip
\noindent \textbf{Concise reason decoding}: This mode produces a compact reasoning response rather than a full CoT trace:
\begin{equation}
    y = \text{ConciseReason}(x) = \{\tilde{r}, \hat{a}\}.
\end{equation}
Such behavior is first acquired from pre-training and can be further enhanced by post-training (\eg, RL). 
By reducing the number of generated tokens, concise reasoning achieves faster inference while preserving response accuracy. 

\smallskip
\noindent \textbf{Direct answer decoding}: The simplest decoding mode generates only the final answer: $y = \text{DirectAnswer}(x) = \{\hat{a}\}$. 
This mode minimizes computation cost for decoding, yet lacking explainability behind the answer.

\subsection{Analysis}
\label{sec:analysis}

In the following analysis, we summarize the key finding at the beginning of each part. We first measure the efficiency of CoT model during training and inference. Then we compare its reasoning capability to the pre-trained model which CoT model was initialized from. Finally, we explore token compression to improve the efficiency of CoT model.

\smallskip
\noindent \textbf{Tested Models}: We consider two typical types of models. 
The first is the pre-trained model, such as \textbf{Qwen2.5-VL~\cite{qwen25}}. 
It is a milestone work of multi-model LLM that demonstrate strong performance on video reasoning. 
The second type is the chain-of-thought (CoT) model, such as \textbf{Video-R1~\cite{video-r1}}. It is initialized by the pre-trained Qwen2.5-VL model and then post-trained to learn CoT reasoning. 
It requires extensive annotations of chain-of-thought and trains the model via supervised fine-tuning (SFT) and reinforcement learning (RL) with GRPO. 

\smallskip
\noindent \textbf{Benchmarks}: We evaluate the models on diverse and commonly-used video benchmarks, including 
(1) \textbf{general video benchmarks} (VideoMME~\cite{fu2024video}, MVBench~\cite{li2023mvbench}): they are created to evaluate the reasoning capability of video LLMs; 
(2) \textbf{long video benchmarks} (MLVU~\cite{zhou2024mlvucomprehensivebenchmarkmultitask}, LVBench~\cite{wang2025lvbench}, LongVideoBench~\cite{wu2024longvideobench}, Egoschema~\cite{mangalam2023egoschemadiagnosticbenchmarklongform}): they are designed for long video understanding; 
(3) \textbf{complex video benchmarks} (VideoHolmes~\cite{cheng2025video}, Video-TT~\cite{video-tt}, MMVU~\cite{zhao2025mmvu}): they emphasize complex event reasoning or multi-discipline understanding.

\smallskip
\smallskip
\infobox{Finding 1: CoT model incurs substantial computation overhead during both training and inference, yet yields modest improvements over the pre-trained model which it was initialized from.}

Fig.~\ref{fig:overhead} (a) reports the training cost of a CoT model. To imitate human-like reasoning, post-training requires \textbf{over 30 hours} on four high-end GPUs, not to mention the additional expense for collecting CoT annotations. 
Further, as shown in Fig.~\ref{fig:overhead} (b), CoT model spends around \textbf{ten times} in inference runtime due to long sequence decoding, when compared to direct answer and concise reason (\eg, 11.9 vs. 0.47 and 1.71). Such high cost hinders the deployment of video LLMs, especially in the resource-constrained environments (\eg, robots). 
By a closer inspection at the generated text, there are frequent self-reflection or verification expressions, such as ``Hmm,'' and ``Wait,''. We hypothesize that these patters are not necessary for \textbf{general video understanding}, which typically require perception and reasoning on the objects, human-object interactions, and social events over time. 

More importantly, despite its significant computation cost, CoT model fails to exhibit benefits on video benchmarks accordingly. 
As shown in Table~\ref{tab:inference_results}, the pre-trained model, which the CoT model was initialized from, can perform favorably to CoT model and event outperforms it on several benchmarks (\eg, VideoMME and EgoSchema). 
Together, these results intrigue our re-thinking about CoT reasoning and its post-training paradigm.

\smallskip
\smallskip
\infobox{Finding 2: Concise reasoning, learned from pre-training, can largely reduce decoding cost, yet its performance is unsatisfactory.}
In Table~\ref{tab:inference_results}, we notice that the pre-trained model already possesses the capability to provide concise reasoning. 
Such reasoning can save substantial computation for decoding.
However, there is a clear performance gap between concise reason and direct answer. 
We conjecture that the pre-trained model is not well aligned with concise reasoning paradigm and could be aligned by post-training.

\smallskip
\smallskip
\infobox{Finding 3: Training-free token compression method can reduce prefilling overhead, but fails to well preserve the performance of concise reasoning.}

In parallel to shorten the text length at decoding phrase, one alternative to improve the efficiency is reducing the overhead at prefilling phrase. 
We explore the recent advances from training-free token compression which do not require further training. 
For example, ~\cite{zhong2024aim} reduces prefilling workload substantially during inference, with minimum loss of pre-trained capability. 
We apply it to the pre-trained and CoT models in a plug-and-play manner.

The results are shown in Table~\ref{tab:token_compression}. Concise reason mode experiences larger accuracy drops, whereas direct-answer or CoT models are less affected.
We conjecture that despite the concise reasoning capability acquired from pre-training, the model is not optimized to such decoding mode and thus being more fragile and vulnerable to the unseen input intervention (\eg, token compression).

\smallskip
\smallskip
\remarkbox{Remark: Motivated by these findings, it is essential to design a solution that can enhance the reasoning capability of pre-trained models in an efficient way.}

\section{Methods}
\label{sec:method}

\smallskip
\smallskip
\remarkbox{Our Hypothesis: Concise reasoning, combined with compressed visual tokens and optimized via RL fine-tuning, is sufficient for effective video reasoning.}
\smallskip
\smallskip

Motivated by our empirical findings, we synthesize an efficient post-training framework tailored for video reasoning. 
As illustrated in Figure~\ref{fig:method}, our method differs from existing approaches in three aspects: (1) \textbf{Post-Training}: it directly performs RL post-training, eliminates the need for SFT and its CoT annotations, (2) \textbf{Prefilling}: it integrates token compression to reduce the cost of training and inference, and (3) \textbf{Decoding}: it generates native, concise reasoning response instead of long CoT text during inference. 
We elaborate each component in the following sections.

\begin{table*}[t]
\centering
\setlength{\tabcolsep}{8pt}
\renewcommand{\arraystretch}{1.4}
\resizebox{0.99\textwidth}{!}
{%
\begin{tabular}{@{}l|c|cc|cccc|ccc@{}}
    \toprule
    \multirow{2}{*}{\textbf{Model}} & 
    \multirow{2}{*}{\textbf{Decoding Mode}} &
    \multicolumn{2}{c|}{\textbf{\footnotesize{General Video Benchmarks}}} & 
    \multicolumn{4}{c|}{\textbf{\footnotesize{Long Video Benchmarks}}} & 
    \multicolumn{3}{c}{\textbf{\footnotesize{Complex Video Benchmarks}}} \\ 
    \cmidrule(l){3-11} 
     &  & {\textbf{\footnotesize{VideoMME}}} & {\textbf{\footnotesize{MVBench}}} & {\textbf{\footnotesize{MLVU}}} & {\textbf{\footnotesize{LVBench}}} & {\textbf{\footnotesize{LongVideoBench}}} & {\textbf{\footnotesize{EgoSchema}}} & {\textbf{\footnotesize{VideoHolmes}}} & {\textbf{\footnotesize{Video-TT}}} & {\textbf{\footnotesize{MMVU}}} \\ \midrule
    Qwen2.5-VL-7B~\cite{qwen25} &  direct answer &  55.5 & 63.2 & 55.4  & 34.2  & 52.5  &  52.5 & 35.7 & 35.2 & 63.4 \\
    Qwen2.5-VL-7B + GRPO~\cite{guo2025deepseek} &  direct answer  & 56.9 (\textcolor{oai-green}{+1.4}) & 64.0 (\textcolor{oai-green}{+0.8}) & 57.7 (\textcolor{oai-green}{+2.3}) & 36.9 (\textcolor{oai-green}{+2.7}) & 54.3 (\textcolor{oai-green}{+1.8}) & 55.7 (\textcolor{oai-green}{+3.2}) & 38.6 (\textcolor{oai-green}{+2.9}) & 38.5 (\textcolor{oai-green}{+3.3}) & 63.7 (\textcolor{oai-green}{+0.3})  \\ \midrule
    Qwen2.5-VL-7B~\cite{qwen25} &  concise reason &  52.8 & 55.2 & 47.3  & 32.2  & 47.5  &  53.3 & 32.1 & 31.4 & 55.2 \\ 
    Qwen2.5-VL-7B + GRPO~\cite{guo2025deepseek} &  concise reason  & 55.4 (\textcolor{oai-green}{+2.6}) & 65.2 (\textcolor{oai-green}{+10.0})   & 58.4 (\textcolor{oai-green}{+11.1})  &  36.0 (\textcolor{oai-green}{+3.8})  & 54.4 (\textcolor{oai-green}{+6.9}) &  53.1 (\textcolor{oai-magenta}{-0.2})  & 40.0 (\textcolor{oai-green}{+7.9}) &  40.1 (\textcolor{oai-green}{+8.7}) & 65.0 (\textcolor{oai-green}{+9.8}) \\ \midrule
    \textcolor{gray}{Qwen2.5-VL-7B + Video-R1~\cite{video-r1}} & \textcolor{gray}{chain-of-thought} & \textcolor{gray}{54.9} & \textcolor{gray}{64.9} & \textcolor{gray}{58.9} & \textcolor{gray}{35.4} & \textcolor{gray}{54.6} & \textcolor{gray}{47.6} & \textcolor{gray}{39.4} & \textcolor{gray}{39.9} & \textcolor{gray}{62.4} \\  \bottomrule
    \end{tabular}%
}
\vspace{-6pt}
\caption{\textbf{Validation 1: Direct reinforcement learning with GRPO} consistently improves pre-trained model and fill the gap of concise reason, without the need for CoT annotations and CoT supervised fine-tuning. 
}
\label{tab:direct_grpo}
\vspace{-6pt}
\end{table*}

\begin{table*}[t!]
\centering
\setlength{\tabcolsep}{8pt}
\renewcommand{\arraystretch}{1.4}
\resizebox{0.99\textwidth}{!}
{%
\begin{tabular}{@{}l|c|cc|cccc|ccc|c@{}}
    \toprule
    \multirow{2}{*}{\textbf{Model}} & 
    \textbf{Decoding} &
    \multicolumn{2}{c|}{\textbf{\footnotesize{General Video Benchmarks}}} & 
    \multicolumn{4}{c|}{\textbf{\footnotesize{Long Video Benchmarks}}} & 
    \multicolumn{3}{c|}{\textbf{\footnotesize{Complex Video Benchmarks}}} & \multirow{2}{*}{\textbf{Avg.}} \\ 
    \cmidrule(l){3-11} 
     & \textbf{Mode} & {\textbf{\footnotesize{VideoMME}}} & {\textbf{\footnotesize{MVBench}}} & {\textbf{\footnotesize{MLVU}}} & {\textbf{\footnotesize{LVBench}}} & {\textbf{\footnotesize{LongVideoBench}}} & {\textbf{\footnotesize{EgoSchema}}} & {\textbf{\footnotesize{VideoHolmes}}} & {\textbf{\footnotesize{Video-TT}}} &  {\textbf{\footnotesize{MMVU}}} & \\ \midrule
    Qwen2.5-VL-7B~\cite{qwen25} & concise reason &  46.7 (\textcolor{oai-magenta}{-6.1}) & 52.0 (\textcolor{oai-magenta}{-3.2}) & 39.4 (\textcolor{oai-magenta}{-7.9})  &  29.5 (\textcolor{oai-magenta}{-2.7}) & 44.4 (\textcolor{oai-magenta}{-3.1}) & 50.7 (\textcolor{oai-magenta}{-2.6}) & 26.9 (\textcolor{oai-magenta}{-5.2}) & 30.2 (\textcolor{oai-magenta}{-1.2}) & 48.2 (\textcolor{oai-magenta}{-7.0}) & \textcolor{oai-magenta}{\textbf{-4.3}} \\
    Qwen2.5-VL-7B + GRPO~\cite{guo2025deepseek} & concise reason &  51.8 (\textcolor{oai-magenta}{-3.6}) & 61.2 (\textcolor{oai-magenta}{-4.0}) & 57.0 (\textcolor{oai-magenta}{-1.4})  &  33.8 (\textcolor{oai-magenta}{-2.2}) & 50.5 (\textcolor{oai-magenta}{-3.9}) & 50.5 (\textcolor{oai-magenta}{-2.6}) & 39.0 (\textcolor{oai-magenta}{-1.0}) & 40.1 (\textcolor{oai-magenta}{-0.0}) & 64.0 (\textcolor{oai-magenta}{-1.0}) & \textcolor{oai-magenta}{\textbf{-2.2}} \\
    \bottomrule
    \end{tabular}%
}
\vspace{-6pt}
\caption{\textbf{Validation 2: Direct reinforcement learning with GRPO} can mitigate the accuracy gap introduced by token compression during inference. The delta (colored) represents the accuracy degradation compared to corresponding model without using token compression. 
}
\label{tab:token_compression_grpo}
\vspace{-8pt}
\end{table*}

\subsection{Post-Training with GRPO}
\label{sec:grpo}

Given a pre-trained MLLM, most prior works (e.g., Video-R1~\cite{video-r1}) adopt a two-stage pipeline that first applies SFT on annotated CoT responses, followed by RL fine-tuning. However, we hypothesize that directly fine-tuning the model with RL is sufficient to enhance reasoning ability and achieve strong performance. Thus, we bypass the costly SFT stage and train the model solely with RL.  

We employ Group Relative Policy Optimization (GRPO)~\cite{shao2024deepseekmath} during RL training, a variant of policy gradient methods. By comparing groups of sampled responses, GRPO removes the need for a critic model and reduces training overhead.  
Formally, given an input $x$ (a video and a textual query), GRPO samples $G$ candidate responses $y=\{y_1, \dots,y_G\}$ from the policy. A reward function assigns scores $\{R_1, \dots, R_G\}$, which are normalized by their mean and standard deviation:
\begin{equation}
\label{eq:ro}
\small
    A_i=
    \frac{R_i-\mathrm{mean}(\{R_i\}_{i=1}^G)}{\mathrm{std}(\{R_i\}_{i=1}^G)} \text{,}
\end{equation}
where $A_i$ denotes the relative advantage of the $i$-th response. This encourages the model to favor higher-scoring responses within the group. In our implementation, we adopt the same reward functions used in most existing works, including \textbf{format reward} and \textbf{accuracy reward}. 
To prevent large deviations from the pre-trained parameters $\pi_\mathrm{ref}$, a KL-divergence term is applied to regularize the model weights. The final optimization objective is:
\begin{equation}
\resizebox{0.40\textwidth}{!}{$
\begin{aligned}
\max_{\pi_\theta} \mathbb{E}_{y \sim \pi_{\theta_{\text{old}}}} \Big[
&\sum_{i=1}^G \min \Big(A_i \cdot \tfrac{\pi_\theta(y_i)}{\pi_{\theta_{\text{old}}}(y_i)}, \ A_i \cdot \text{clip}(\tfrac{\pi_\theta(y_i)}{\pi_{\theta_{\text{old}}}(y_i)}, \\
& 1-\epsilon, 1+\epsilon) \Big) -\ \beta \, \mathrm{D}_{\text{KL}}(\pi_\theta \parallel \pi_{\text{ref}})
\Big],
\end{aligned}
$} 
\end{equation}
where $\pi_{\theta_{\text{old}}}$ is the old policy before gradient update, $\pi_{\text{ref}}$ denotes the reference policy (\eg, the model before GRPO training), and $\beta$ is a regularization coefficient controlling divergence from the reference policy.

\subsection{Prefill with Token Compression}
\label{sec:token_compression}
Videos often contain redundant visual tokens, significantly increasing the computational cost of LLM prefilling and response decoding. 
Inspired by the recent method AIM~\cite{zhong2024aim}, we perform token merging and token pruning to reduce the number of visual tokens. 
Unlike AIM designed for inference, we incorporate token compression to training and let the model learn to adapt the compressed information.

Existing token compression methods, including AIM, typically do not physically remove tokens from the model. 
Instead, they manipulate attention masks to indicate the retained or pruned tokens. It is easy to implement, yet leaves both wall time and GPU memory consumption unchanged and still large.
In contrast, to support efficient training, we develop the token pruning that physically discards unimportant tokens to reduce memory footprint and runtime.

Moreover, many existing token compression methods require explicit computation of attention weights and thus are not compatible with FlashAttention~\cite{dao2023flashattention}.
To handle this incompatibility, we selectively disable FlashAttention in only a small number of layers that perform pruning (\eg, 5 layers). 
This hybrid strategy preserves the efficiency benefits from both FlashAttention and token compression. Hence, our final implementation achieves substantial memory and runtime savings during both training and inference.

Once token compression is applied, the number of tokens and LLM computation are largely reduced. This allows the model to take more video frames as inputs and thus better understand long videos, using comparable computation as the original network forward without token compression.

\subsection{Decode with Concise Reasoning Mode}

As introduced in Section~\ref{sec:background}, we adopt the decoding mode as concise reasoning. It provides explainability to the final answer and meanwhile maintains a reasonable decoding overhead, making both training and inference efficient. 

\begin{table*}[t!]
\centering
\setlength{\tabcolsep}{8pt}
\renewcommand{\arraystretch}{1.4}
\resizebox{0.98\textwidth}{!}
{%
\begin{tabular}{@{}l|cc|cc|cccc|ccc@{}}
    \toprule
    \multirow{2}{*}{\textbf{Model}} & 
    \textbf{Training} & 
    \textbf{Decoding} &
    \multicolumn{2}{c|}{\textbf{\footnotesize{General Video Benchmarks}}} & 
    \multicolumn{4}{c|}{\textbf{\footnotesize{Long Video Benchmarks}}} & 
    \multicolumn{3}{c}{\textbf{\footnotesize{Complex Video Benchmarks}}} \\ 
    \cmidrule(l){4-12} 
    & \textbf{Stages} & \textbf{Mode} & {\textbf{\footnotesize{VideoMME}}} & {\textbf{\footnotesize{MVBench}}} & {\textbf{\footnotesize{MLVU}}} & {\textbf{\footnotesize{LVBench}}} & {\textbf{\footnotesize{LongVideoBench}}} & {\textbf{\footnotesize{EgoSchema}}} & {\textbf{\footnotesize{VideoHolmes}}} & {\textbf{\footnotesize{Video-TT}}} & {\textbf{\footnotesize{MMVU}}} \\ \midrule
    Qwen2.5-VL-7B~\cite{qwen25} & PT & direct answer &  55.5 & 63.2 & 55.4  & 34.2  & 52.5  &  52.5 & 35.7 & 35.2 & 63.4 \\
    Qwen2.5-VL-7B~\cite{qwen25} & PT & concise reason &  52.8 & 55.2 & 47.3  & 32.2  & 47.5  &  53.3 & 32.1 & 31.4 & 55.2 \\
    Video-R1-7B~\cite{video-r1} & PT, SFT, RL & chain-of-thought & 54.9 & 64.9 & 58.9 & 35.4 & 54.6 & 47.6 & 39.4 & 39.9 & 62.4 \\  \midrule
    Our Final Model & PT, RL & concise reason  & 60.6 & 65.6   & 67.0  &  38.9  &  55.7 &  55.0 & 41.6 & 40.4 & 63.6 \\ \midrule
    \rowcolor{gray!15} Qwen3-VL-8B~\cite{qwen3} & PT & direct answer & 60.3 & 66.8 & 58.2 & 35.8 & 56.2 & 63.1 & 40.7 & 39.6 & 66.4 \\
    \rowcolor{gray!15} Qwen3-VL-8B~\cite{qwen3} & PT & concise reason & 57.5 & 63.5 & 52.7 & 32.7 & 52.7 & 62.8 & 37.4 & 34.7 & 67.0 \\
    \bottomrule
    \end{tabular}%
}
\vspace{-6pt}
\caption{\textbf{Final model results on video benchmarks}. 
Our final solution performs token compression at both training and inference, and fine-tunes the pre-trained model Qwen2.5-VL-7B with concise reasoning mode using GRPO. 
Ours not only achieves the best results across benchmarks, but also requires much less computation during training (\ie, no SFT) and inference (\ie, no heavy CoT).}
\label{tab:video_bench}
\vspace{-6pt}
\end{table*}

\begin{table*}[t]
\centering
\setlength{\tabcolsep}{8pt}
\renewcommand{\arraystretch}{1.4}
\resizebox{0.99\textwidth}{!}
{%
\begin{tabular}{@{}l|c|cc|cccc|ccc@{}}
    \toprule
    \multirow{2}{*}{\textbf{Model}} & 
    \textbf{Decoding} &
    \multicolumn{2}{c|}{\textbf{\footnotesize{General Video Benchmarks}}} & 
    \multicolumn{4}{c|}{\textbf{\footnotesize{Long Video Benchmarks}}} & 
    \multicolumn{3}{c}{\textbf{\footnotesize{Complex Video Benchmarks}}} \\ 
    \cmidrule(l){3-11} 
     & \textbf{Mode} & {\textbf{\footnotesize{VideoMME}}} & {\textbf{\footnotesize{MVBench}}} & {\textbf{\footnotesize{MLVU}}} & {\textbf{\footnotesize{LVBench}}} & {\textbf{\footnotesize{LongVideoBench}}} & {\textbf{\footnotesize{EgoSchema}}} & {\textbf{\footnotesize{VideoHolmes}}} & {\textbf{\footnotesize{Video-TT}}} &  {\textbf{\footnotesize{MMVU}}} \\ \midrule
    Qwen2.5-VL-7B + GRPO~\cite{guo2025deepseek} & concise reason  & 55.4 & 65.2   & 58.4  &  36.0  &  54.4 &  53.1  & 40.0 & 40.1 & 65.0 \\ \midrule
    inference w. T.C. & concise reason  & 51.8 &  61.2 & 57.0 & 33.8  & 50.5 & 50.5 & 39.0 & 40.1 & 64.0 \\
    inference w. T.C. \& 6x frames & concise reason  & 59.5 &  65.2  & 62.9  &  38.0  &  54.9 &  54.8 & 40.8 & 39.9 & 65.3 \\
    + training w. T.C. \& 6x frames & concise reason  & 60.6 & 65.6   & 67.0  &  38.9  &  55.7 &  55.0 & 41.6 & 40.4 & 63.6 \\
    \bottomrule
    \end{tabular}%
}
\vspace{-6pt}
\caption{\textbf{Ablation study on token compression and the number of video frames}. T. C. denotes token compression method.}
\label{tab:ablation}
\vspace{-10pt}
\end{table*}

\section{Experiments and Results}

In this section, we start with the experiments validating our hypothesis and then present the results of our final model. Further, we provide ablation study and output visualization.

\subsection{Avoid Overthinking}
\smallskip
\smallskip
\infobox{Direct GRPO training consistently enhances the pre-trained model and bridges the gap in concise reasoning, without requiring CoT annotations or SFT.}

We directly fine-tune the pre-trained model through RL with GRPO algorithm. The results are reported at Table~\ref{tab:direct_grpo}. 
It turns out that consistent improvements are surprisingly observed across benchmarks (\eg, +1.4/+2.6 on VideoMME, +2.3/+11.1 on MLVU, +2.9/+7.9 on VideoHolmes, over direct answer/concise reason mode).
Further, the GRPO fine-tuned model even surpasses the heavily post-trained CoT model (\eg, 55.4 vs.~54.9 on VideoMME, 40.0 vs.~39.4 on VideoHolmes). 
These consistent improvements suggest that GRPO can generally incentive the reasoning capability, \textbf{eliminating the need for CoT annotations or CoT SFT}.

In Table~\ref{tab:direct_grpo}, we also observe that, for the pre-trained model, there exists a noticeable performance gap between direct answer and concise reason modes (\eg, 2.7 on VideoMMME, 8.1 on MLVU). 
We conjecture that this is caused by the pre-training data which emphasizes on answer accuracy rather than explicit reasoning. 
Interestingly, \textbf{after GRPO fine-tuning, this gap is mitigated}. Concise reason mode can even perform better (\eg, 65.2 vs. 64.0 on MVBench, 58.4 vs. 57.7 on MLVU, 40.0 vs. 38.6 on VideoHolmes).
These results validate our hypothesis that concise reasoning, optimized by only RL fine-tuning without overthinking, can be sufficient for video reasoning.

\subsection{Look Less, See Further}
\smallskip
\smallskip
\infobox{Direct GRPO training improves the compatibility of concise reasoning with token compression, allowing for more input frames with improved accuracy.}
In Table~\ref{tab:token_compression_grpo}, we observe that concise reasoning can be further integrated with token compression after GRPO fine-tuning (\eg, performance drops shrink from –4.3 to –2.2). 
This in turn allows us to input more frames for improved performance, which is discussed in our Ablation Study (Sec.~\ref{sec:ablation}).
We conjecture that GRPO training strengthens the model’s behavior for concise reasoning, making it more capable of operating effectively in this decoding mode. 
The degree to which the model acquires this capability could directly influence its robustness to unseen input perturbations, such as token compression.

\begin{figure*}[t]
    \centering
    \includegraphics[width=0.95\linewidth]{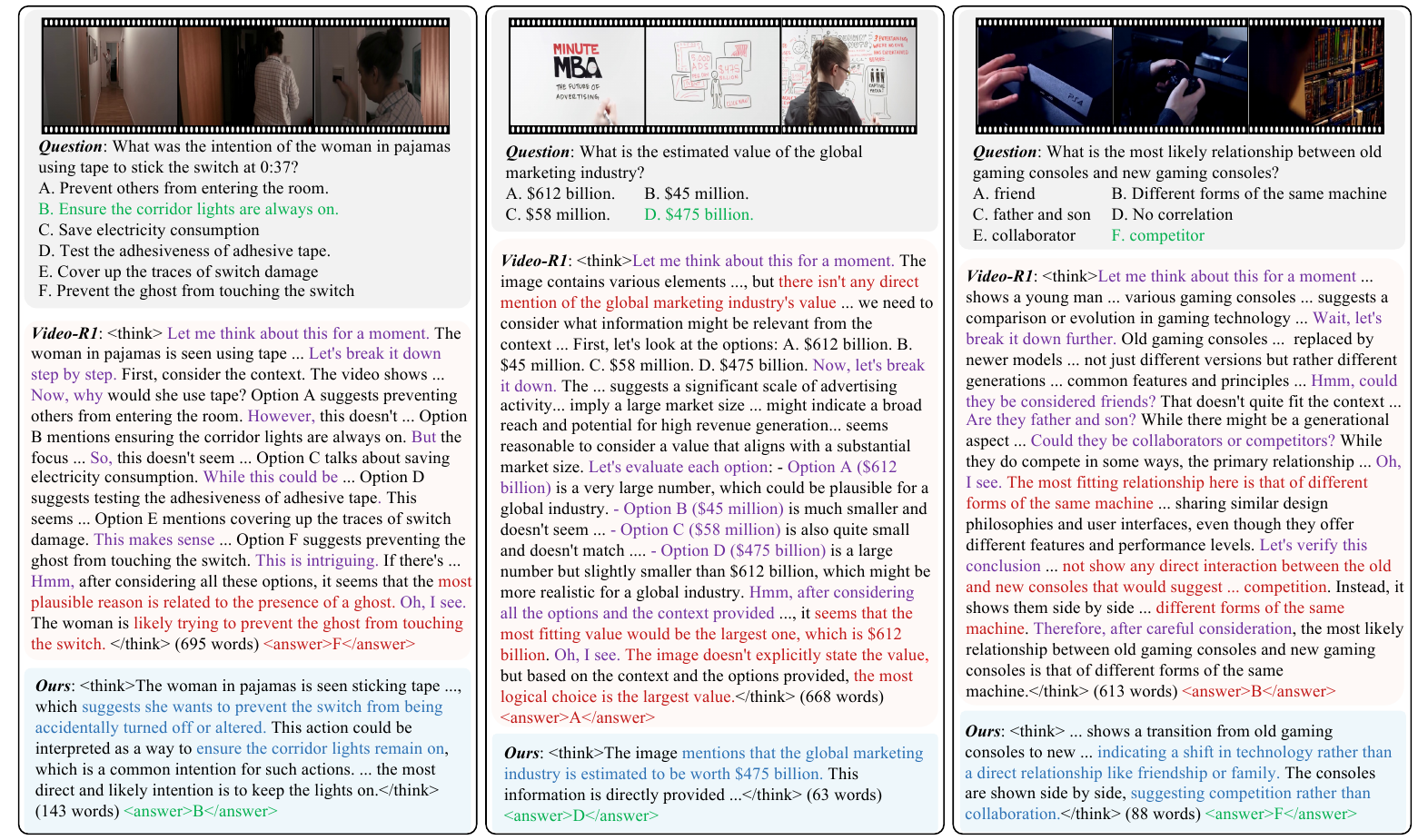}
    \vspace{-8pt}
    \caption{\textbf{Visualization of generated text} from CoT reasoning (Video-R1) and concise reasoning (Ours). Partial text chunks are colored through human validation.
    \textcolor{visualgreen}{Green}: ground-truth or correct predicted answers.
    \textcolor{visualblue}{Blue}: correct intermediate reasoning steps.
    \textcolor{visualpurple}{Purple}: unnecessary intermediate reasoning steps.
    \textcolor{visualred}{Red}: incorrect intermediate reasoning steps or final predictions.
    } 
    \vspace{-10pt}
    \label{fig:visualization}
\end{figure*}

\subsection{Results of Final Model}

\smallskip
\noindent \textbf{Implementation Details}: 
We use the same training dataset and training hyper-parameters as Video-R1~\cite{video-r1}. Specifically, we train models via GRPO for 2,000 steps on Video-R1-260k. For efficiency consideration, during both training and inference, we keep each frame with resolution as $128 \times 28 \times 28$ and each video with the maximum number of frames as 16, unless otherwise noted.

\smallskip
\noindent \textbf{Setup}: We compare the performance of our final model with multiple baselines, including pre-trained Qwen2.5-VL~\cite{qwen25} with direct answer or concise reason mode, and CoT model Video-R1~\cite{video-r1}. 
For a fair comparison, we maintain the same level of computation budget for prefilling. Our final model thus takes more input frames during training and inference (\ie, 96 frames), since token compression technique reduces the number of visual tokens for each frame.

\smallskip
\noindent \textbf{Our model achieves strong results across benchmarks}. As shown in Table~\ref{tab:video_bench}, our model achieves the best results on all benchmarks (\ie, +5.7 on VideoMME, +8.1 on MLVU, +2.2 on VideoHolmes, over Video-R1). 
Notably, using only the resources for GRPO fine-tuning, our model can even outperform or compare favorably to Qwen3-VL which takes much more resources during pre-training to improve beyond Qwen2.5-VL.
These results suggest that our method can effectively enhance the reasoning capability on general videos, long videos, and complex videos.

\smallskip
\noindent \textbf{Our method is efficient at both training and inference}. 
For training, our method only requires RL post-training by inheriting the concise reasoning capability derived from pre-training. It eliminates the need for CoT annotations and CoT SFT.
For inference, our method performs concise reasoning rather than generating long CoT traces.

\subsection{Ablation Study}
\label{sec:ablation}
\smallskip
As shown in Table~\ref{tab:ablation}, incorporating token compression during training yields clear performance gains compared to applying it only at inference time (e.g., +1.1 on VideoMME, +4.1 on MLVU, and +0.8 on VideoHolmes).
These results indicate that the model can learn to adapt to compressed visual tokens during training.

Moreover, token compression reduces the number of visual tokens, allowing the model to process more video frames within a comparable prefilling computation budget.
This expanded temporal coverage significantly enhances performance, particularly for long video understanding tasks (e.g., +5.9 on MLVU and +4.2 on LVBench).

\subsection{Visualization}
Figure~\ref{fig:visualization} presents qualitative comparisons between CoT reasoning (Video-R1) and concise reasoning (our final model). Given a video, a question, and multiple-choice options, the model is instructed to generate the reasoning process / final anser within the ``think'' / ``answer'' brackets.

\smallskip
\noindent \textbf{Long reasoning traces are often unnecessary and even distracting.} Across the three examples, CoT reasoning tends to overthink and overcomplicate the problem, for instance, elaborating on each option with human-like phrases such as ``Oh, I see'' or ``Hmm''.
These expressions are format-oriented rather than content-oriented, indicating that the model focuses more on mimicking human reasoning style than on actual reasoning content.
Such verbose reasoning not only increases computational cost, but can also divert the reasoning trajectory, leading to incorrect answers.

\smallskip
\noindent \textbf{Dense frames help.} 
With token compression, the model can process densely sampled video frames without increasing prefilling overhead. Dense frame coverage preserves critical visual details that may be missed under sparse sampling.
Take the middle sample in Figure~\ref{fig:visualization} as the instance. A key frame directly reveals the correct answer. This frame is missed by sparse sampling and CoT reasoning is unable to compensate such missing information.

\section{Conclusion}
In this work, we revisit the role of chain-of-thought reasoning in video MLLMs and challenge the assumption that long, human-like reasoning traces and large sets of video tokens are necessary for strong performance. 
Through extensive benchmarking, we find that conventional CoT provides limited benefits while incurring significant computational overhead, and that pre-trained models are poorly aligned with concise reasoning despite possessing sufficient knowledge. 
Motivated by these observations, we hypothesize and empirically validate that effective video reasoning can emerge from concise reasoning traces operating on compressed video tokens. 
Our post-training and inference framework realizes this paradigm, substantially improving inference efficiency while maintaining competitive accuracy across diverse video understanding tasks, without reliance on CoT annotations or heavy SFT. 
We hope these findings inspire a rethinking of how reasoning should be modeled in MLLMs and encourage future exploration of efficient approaches to video reasoning.

{
    \small
    \bibliographystyle{ieeenat_fullname}
    \bibliography{main}
}


\end{document}